\documentclass[10pt,onecolumn,a4paper]{templates/fairmeta}

\usepackage{multirow}
\usepackage{tabularx}
\usepackage{makecell}
\usepackage{stmaryrd}

\usepackage{xspace}
\usepackage{amsmath,amsfonts,bm,amsthm}
\usepackage{amssymb}
\usepackage{mathtools}
\usepackage{sidecap}
\usepackage{enumitem}
\usepackage{soul}
\usepackage{array}

\usepackage{caption}
\usepackage{subcaption}
\usepackage{ifthen}
\usepackage{tikz}
\usetikzlibrary{arrows.meta,arrows,tikzmark,positioning,calc}
\usepackage{float}
\usepackage{microtype}
\usepackage{graphicx}

\usepackage{booktabs}
\usepackage{wrapfig}
\usepackage{needspace}
\makeatletter
\newcommand\wrapfill{\par
  \ifx\parshape\WF@fudgeparshape
    \nobreak
    \ifnum\c@WF@wrappedlines>\@ne
      \advance\c@WF@wrappedlines\m@ne
      \vskip\c@WF@wrappedlines\baselineskip
      \global\c@WF@wrappedlines\z@
    \fi
    \allowbreak
    \WF@finale
  \fi}
\makeatother

\usepackage{rotating}
\usepackage{multirow}
\usepackage[table,dvipsnames]{xcolor}

\makeatletter
\DeclareRobustCommand\onedot{\futurelet\@let@token\@onedot}
\def\@onedot{\ifx\@let@token.\else.\null\fi\xspace}
\makeatother

\def\vs{\emph{vs}\onedot}

\def\odec{$\downarrow$}

\def\upsymb#1{\ensuremath{\ifcase#1\or *\or \dagger\or \ddagger\or
   \mathsection\or \mathparagraph\or \|\or **\or \dagger\dagger
   \or \ddagger\ddagger \else\@ctrerr\fi}}

\usepackage{cleveref}
\crefname{section}{Sec.}{Secs.}
\Crefname{section}{Section}{Sections}
\crefname{table}{Tab.}{Tabs.}
\Crefname{table}{Table}{Tables}
\crefname{figure}{Fig.}{Figs.}
\Crefname{figure}{Figure}{Figures}
\crefname{appendix}{App.}{App.}
\Crefname{appendix}{Appendix}{App.}
\crefname{equation}{Eq.}{Eqs.}

\usepackage{marginnote}

\def\mypar#1{\vspace{0mm}\noindent\textbf{#1.}\hspace{0mm}}
\newcolumntype{H}{>{\setbox0=\hbox\bgroup}c<{\egroup}@{}}

\def\tabfigendspace{{\vspace{0mm}}}
\def\figmidspace{{\vspace{0mm}}}

\newcommand{\cs}{\mathcal{S}}

\newcommand{\jit}[1][]{\textsc{JiT\ifthenelse{\equal{#1}{}}{}{-#1}}\xspace}
\newcommand{\ujit}[1][]{\textsc{U-JiT\ifthenelse{\equal{#1}{}}{}{-#1}}\xspace}
\newcommand{\ljit}[1][]{\textsc{$l$-JiT\ifthenelse{\equal{#1}{}}{}{-#1}}\xspace}
\newcommand{\lcosmos}[1][]{\ljit[#1] (Cosmos Tokenizer)}
\newcommand{\lwanone}[1][]{\ljit[#1] (Wan2.1 Tokenizer)}
\newcommand{\lwantwo}[1][]{\ljit[#1] (Wan2.2 Tokenizer)}
\newcommand{\lltx}[1][]{\ljit[#1] (LTX Tokenizer)}
\def\oursampling{}
\def\basesamp{(base sampling)}

\definecolor{cbetter}{HTML}{12a13f}
\definecolor{cworse}{HTML}{d62613}

\newcommand{\vworse}[1]{{\itshape\bfseries\color{cworse}#1}}
\newcommand{\vbetterl}[1]{{\itshape\color{cbetter!55!gray}#1}}
\newcommand{\vworsel}[1]{{\itshape\color{cworse!55!gray}#1}}
\newcommand{\vsame}[1]{{\itshape#1}}

\definecolor{cyellow}{HTML}{fff2cc}
\definecolor{cblue}{HTML}{cfe2f3}
\definecolor{cgreen}{HTML}{d9ead3}
\definecolor{cpink}{HTML}{f4d4e2}
\definecolor{cpurple}{HTML}{bbb0d4}
\definecolor{corange}{HTML}{fde4cd}
\definecolor{cred}{HTML}{e6b8af}

\newcommand{\papertitle}{Pixel-space video generation with U-JiT}

\title{\papertitle}
\author[1,2]{Théophane Vallaeys}
\author[1]{Jakob Verbeek}
\author[2,3]{Matthieu Cord}

\affiliation[1]{FAIR, Meta}
\affiliation[2]{Sorbonne University, Paris}
\affiliation[3]{Valeo.ai}

\abstract{Latent diffusion and flow matching models are ubiquitous in state-of-the-art generative video models.
While highly effective,  relying on
pretrained autoencoders to compress the data into a lower-dimensional latent space also comes with drawbacks.
First, training proceeds in two stages: autoencoder pretraining, with compression and reconstruction objectives, followed by latent generative model training, precluding fully end-to-end training.
Second, autoencoders also lead to an inherent loss of signal, and the trade-off between reconstruction quality and generative performance in the latent space is still an active area of research.
In this paper we explore the viability of generative video models directly in pixel space.
We analyze recent approaches in pixel-space generative modeling of images that rely on basic transformer designs, and identify several shortcomings that limit their effectiveness for the high-dimensional pixel-space video generation.
To overcome these limitations, we propose a UNet-like transformer  architecture along with improvements to the  training recipe and sampling strategy.
We conduct controlled experiments, training latent and pixel-space models from scratch on Kinetics-600.
For the standard frames-to-video task at 128 resolution, our \ujit model improves generative FVD by 27\% over prior  state-of-the-art results, and by 14\% against the JiT pixel-space baseline.
Scaling to higher 256 resolution, we find the improvement over pixel-space and latent-based models grows as we increase the  spatiotemporal compression rate up to $64\!\times\!64\!\times\!8$. Finally, we  find that \ujit improves per-frame FID by 27\% over latent models and 61\% over pixel-space models on  class-to-video generation, with respectively 9\% and 56\% reduction in FVD.
}

\date{October 29, 2026}
\correspondence{Théophane Vallaeys (webalorn@meta.com)}

\begin{document}

\maketitle

\begin{figure}[h]
    \centering
    \scalebox{0.88}{\input{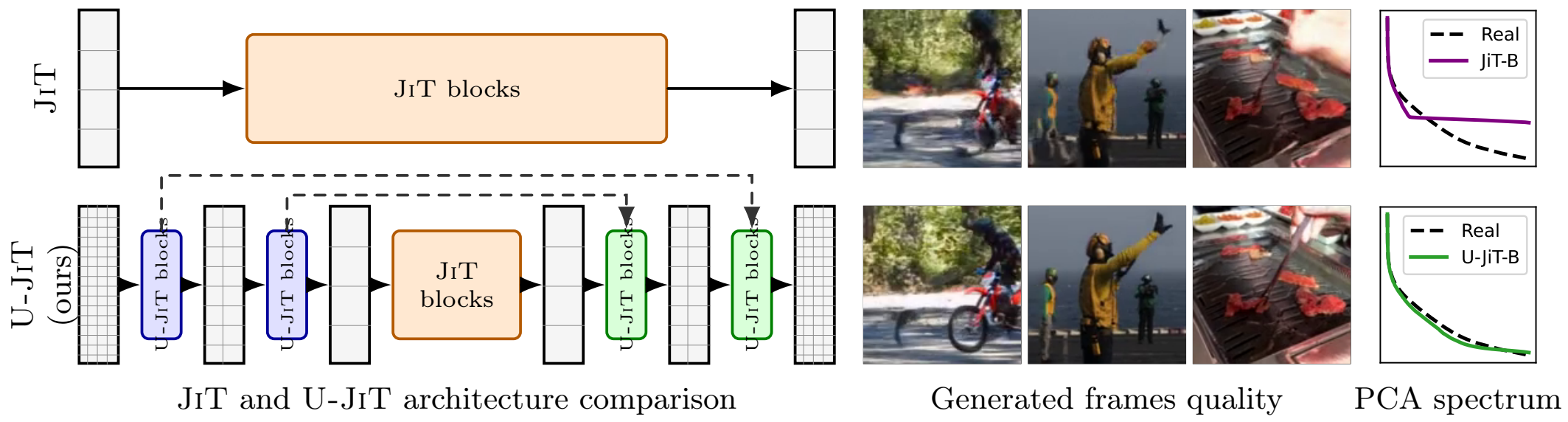}}
    \caption{
        {\bf Comparison of \jit{} and our proposed \ujit{} on video generation.}
        While \jit{} processes the video with a single stack of large-patch blocks, \ujit{} varies the patch size along model depth.
        In this manner the model benefits from small patches in the outer layers to capture high resolution detail, as well as high channel depth in the inner layers to model high-level semantic structure.
        On the right, frames generated by each model, and the PCA spectrum of generated \vs real videos.
    }
    \label{fig:teaser}
    \vspace{-1em}
\end{figure}

\section{Introduction}
\label{sec:intro}

Recent advances on pixel-space image generation~\citep{Hoogeboom_2025_CVPR,chen2026asymmetricflowmodels,chen2026l2punlockinglatentpotential,pavasovic2026waitsignalsimplefrequencyaware,baade2026latent} have shown that Flow-Matching models can achieve high quality generation in ambient space, without relying on the latent spaces generated by autoencoders or ``tokenizers''.
By using transformer-based models with large patch size, they also preserve the scalability and efficiency of their latent counterparts.
Flow-Matching models are trained to predict, from a noised input $x_t=(1-t)\varepsilon + t x_0$, the velocity $v_t = x_0 - \varepsilon$ that transports the distribution of noise $\varepsilon$ to the distribution of data $x_0$.
The key idea underlying  pixel-space models is the hypothesis that the natural data lies on a low-dimensional manifold embedded in the high-dimensional ambient space~\citep{turk91jcn,tipping99neco,roweis00science, carlsson2009topology, chapelle2006semisupervised}.
This allows the predicted velocity to be decomposed into a compressible signal $x_0$ and the incompressible Gaussian noise $\varepsilon$, either explicitly through the objective~\citep{jit_Li_2026_CVPR}, or implicitly through the architecture design~\citep{pixeldit_Yu_2026_CVPR,Ma_2026_CVPR,Chen_2026_CVPR}.

However, state-of-the-art video generation models~\citep{wan2025wanopenadvancedlargescale,polyak2025moviegencastmedia,ICLR2025_3ab228c4,ICLR2025_ce31378e,walt} rely on highly compressed latent representations~\citep{hacohen2024ltxvideorealtimevideolatent,fsvideoteam2026fsvideofastspeedvideo,chen2025dcvideogenefficientvideogeneration} to achieve fast generation at scale, taking advantage of the high  redundancy in  video signal.
Direct application of pixel-space models for images to video is complicated by the additional temporal dimension.
Simple linear patchification layers cannot properly compress the high-dimensional data, as per-patch dimension becomes significantly higher than the model hidden dimension.

In our work, we propose to address this limitation directly at the architectural level.
After demonstrating how plain transformers struggle to generate high frequencies in video, we introduce \ujit, a minimal adaptation of the transformer architecture based on U-shaped transformers~\citep{udit_NEURIPS2024_5d2e24df}, see \Cref{fig:teaser}.
We propose to keep the \textit{stack of transformer blocks} as the main backbone, but to vary the patch size along the depth of the network, using single-pixel patches at the finest level.
This design allows fine-grained modeling of high-dimensional data at the finest level coupled with a learned, progressive, non-linear compression of the pixel-space video to the internal hidden representations.
Skip connections are added to allow propagation of high-frequency information, and adaptations to the attention mechanism in the early and late blocks are used to preserve efficiency.

We evaluate our method on the Kinetics-600~\citep{carreira2018shortnotekinetics600} dataset, and show on the frames-to-video task that \ujit improves FVD and FID by respectively 29\% and 90\% over the \jit pixel-space baseline, outperforms by 27\% the FVD of the best published results on the benchmark, and achieves competitive FVD with 64\% reduction in FID compared to our best latent-space baseline.
We additionally show that among those models, \ujit scales the best with respect to the patch size, supporting strong spatiotemporal compression that \jit models struggle with.
Finally, for the more challenging class-conditional generation setting, we improve FVD by 9\% over the best latent baseline, and by 56\% over the \jit pixel-space baseline, with a high compression rate of f32t8.

We summarize our contributions as follows:
(1) We provide a detailed experimental analysis of the plain transformer architecture for pixel-space video generation, and show its practical limitations.
(2) We propose \ujit, a simple and efficient adaptation of the transformer architecture for pixel-space video generation, with minimal changes to preserve efficiency and scalability.
(3) We study three key behaviors of two-stage latent-space video generation models, and show how standard Flow-Matching mechanisms can be used as their replacements, improving the results of both \jit and \ujit.
(4) We conduct extensive experiments across model sizes and tasks on Kinetics-600, showing that \ujit outperforms existing latent-space and pixel-space baselines across both FVD and FID  metrics.

\section{Related work}
\label{sec:related}

\mypar{Diffusion and Flow-matching}
Diffusion~\citep{pmlr-v37-sohl-dickstein15,NEURIPS2020_4c5bcfec,song2021scorebasedgenerativemodelingstochastic,NEURIPS2021_49ad23d1,song2021denoising,pmlr-v139-nichol21a} is a class of generative process modeling stochastic paths between a source (noise) and a target (data) distribution by learning to reverse a forward noising process.
Flow Matching~\citep{lipman2023flow,liu2023flow,albergo2023building} can be seen as a variant learning a deterministic velocity field between source and target.
Notable design choices of the process itself include noise scheduling~\citep{chen2023importancenoiseschedulingdiffusion}, sampling method~\citep{NEURIPS2022_a98846e9}, guidance mechanism~\citep{ho2021classifierfree,NEURIPS2024_dd540e1c}, training timestep sampling~\citep{esser2024scaling}
and log-SNR shift with data dimensionality~\citep{pmlr-v202-hoogeboom23a}.
Those methods are considered state-of-the-art for image~\citep{Peebles_2023_ICCV,ma2024sit,esser2024scaling} and video~\citep{video_diffusion_ho2022,ho2022imagenvideohighdefinition,singer2022makeavideotexttovideogenerationtextvideo,blattmann2023stablevideodiffusionscaling,ma2025latte,walt,vdt} generation.

\mypar{Latent-space generation}
Latent diffusion models~\citep{Rombach_2022_CVPR} were introduced to allow high-dimensional generation by first compressing the data into a lower-dimensional latent space with an autoencoder~\citep{kingma2022autoencodingvariationalbayes,NIPS2017_7a98af17,Esser_2021_CVPR}.
These models inherently carry a trade-off between compression and generation quality~\citep{ICLR2025_f062da19}.
Alignment with foundation models~\citep{ldit_Yao_2025_CVPR,ICLR2026_fa5ddd6b} and direct use of SSL encoders~\citep{ICLR2026_3c4141c1,singh2026improvedbaselinesrepresentationautoencoders} have been shown to improve the quality of latent generation and to allow larger latent dimensions.
On the other end, diffusion decoders~\citep{zhao2025epsilonvae,chen2025diffusionautoencodersscalableimage,vallaeys2026ssddsinglestepdiffusiondecoder,lu2026pidfasthighresolutionlatent} focus on generating missing frequencies from the latent encoding.
These autoencoders have been successfully extended to
discrete~\citep{yu2024languagemodelbeatsdiffusion} and continuous video tokenization~\citep{guo2025deradecoupledrepresentationalignment,walt,ICLR2025_ce31378e,chen2025dcvideogenefficientvideogeneration,nvidia2025cosmosworldfoundationmodel,polyak2025moviegencastmedia,wan2025wanopenadvancedlargescale,hacohen2024ltxvideorealtimevideolatent,kong2025hunyuanvideosystematicframeworklarge}.
Various improvements have also been proposed for the latent model architecture, such as using and improving a transformer backbone~\citep{Peebles_2023_ICCV,ldit_Yao_2025_CVPR,ddt_Wang_2026_CVPR}, and using U-Net~\citep{ronneberger15miccai} inspired architectures~\citep{10.1007/978-3-031-73242-3_3,Bao_2023_CVPR,udit_NEURIPS2024_5d2e24df,pmlr-v235-crowson24a}.
\looseness=-1

\mypar{End-to-end pixel-space generation}
The main limitation of latent models is the two-stage training, which can introduce a loss of information and a latent space that is hard to diffuse.
One way to address it is to jointly train the tokenizer and the latent model, either as a first-stage improvement~\citep{heek2026unifiedlatentsultrain}, a finetuning stage after pretraining~\citep{Leng_2025_ICCV}, or as a fully end-to-end training~\citep{duggal2026endtoendtrainingunifiedtokenization,chu2026endtoend,meshchaninov2026how}.
Another line of work tries to address the issue by directly generating in pixel space.
Transformers are adapted by using additional convolutional layers~\citep{pmlr-v202-hoogeboom23a,Hoogeboom_2025_CVPR}, pixel-space specialized output heads~\citep{Ma_2026_CVPR,Chen_2026_CVPR,pixeldit_Yu_2026_CVPR}, or by keeping a plain transformer backbone with a reformulation of the target velocity~\citep{jit_Li_2026_CVPR}.
We build upon this line of work, by identifying and overcoming some of the limitations of current pixel-space approaches.
\looseness=-1

\section{Method}
\label{sec:method}

\subsection{The architectural bottleneck on high-dimensional data}
\label{subsec:method_bottleneck}

Natural images and videos lie on low-dimensional manifolds embedded in high-dimensional ambient space, which implies that the intrinsic dimension of the data is much lower than the noise or velocity dimension~\citep{turk91jcn,tipping99neco,roweis00science, carlsson2009topology, chapelle2006semisupervised}.
This has been leveraged in \jit~\citep{jit_Li_2026_CVPR} by predicting the clean image rather than the velocity field directly, this separates the incompressible noise present in the velocity term from the compressible (clean image) signal while maintaining a simple architectural design.
We show here that this is not sufficient to capture the distribution of high-dimensional data such as videos.
The relationship between the low-dimensional manifold and the high-dimensional ambient space is non-linear, implying that standard transformers are insufficient to capture this manifold with large patch sizes,
resulting in lost high frequencies.

\begin{wrapfigure}{l}{0.33\columnwidth}
    \vspace{-\intextsep}
    \centering
    \includegraphics[width=\linewidth]{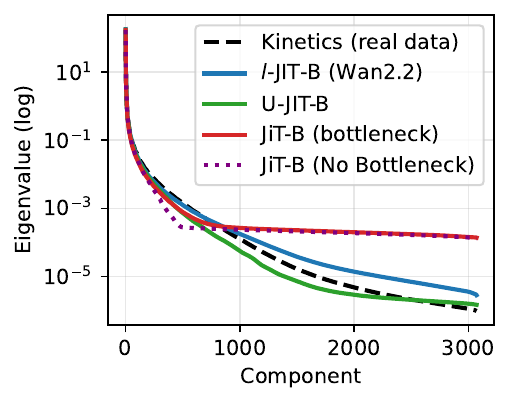}
    \vspace{-2em}
    \caption{
        {\bf PCA spectrum} over $16\times16\times4$ patches, computed over 1k real and sampled videos.
    }
    \label{fig:pca_spectrum}
    \vspace{-1em}
\end{wrapfigure}

To empirically display this behavior on real data, we train a pixel-space \jit[B]{} model on Kinetics-600 for 200 epochs on the standard task of generating 12 frames from 5 starting frames, with a patch size of $16\times16\times4$ and a frame resolution of $256\times256$.
We additionally train a latent-space model (\lwantwo[B]{}) and our custom architecture (\ujit[B]{}, see \Cref{subsec:method_architecture}) on the same task.
In \Cref{fig:pca_spectrum} we display the PCA spectrum of real and generated space-time video patches.
We observe that \jit fails to capture high frequencies, as there is a sharp cut between two regimes of the spectrum: the first one roughly follows a power law, similarly to real data, while the second one is close to a flat Gaussian noise spectrum.\footnote{
By removing the input bottleneck projection, we confirmed that this frequency cut is related to the inner feature dimension of the model rather than the input bottleneck.}
On the other hand, both the latent model and our  \ujit[B] model manage to capture the power law structure of the real data spectrum across the full  range of frequencies.

To confirm this observation, we propose to model the distribution of real and generated patches as a mixture of two signals: a lower-dimensional signal $P$ that captures the compressible part of the data (the signal subspace), and random Gaussian noise $\varepsilon \sim \mathcal{N}(0, \sigma^2 I)$.
This model is well aligned with the Flow-Matching generation process, which shifts samples from pure noise to data.
We follow \citet{gavish2014optimal} to empirically estimate the threshold between signal and noise in the PCA spectrum\footnote{We follow Eq.~26, Section~E, to empirically estimate the noise level from the data.}.
We find an estimated signal-subspace dimension of 1308 for real Kinetics-600 videos, 1258 for videos generated in the Wan2.2 tokenizer latent space, 1284 for our \ujit{}-generated videos, and only 596 for \jit-generated videos.
This confirms that \jit fails to capture a large part of the signal, while the \ujit architecture allows a better split between signal and noise, capturing a much wider range of frequencies.

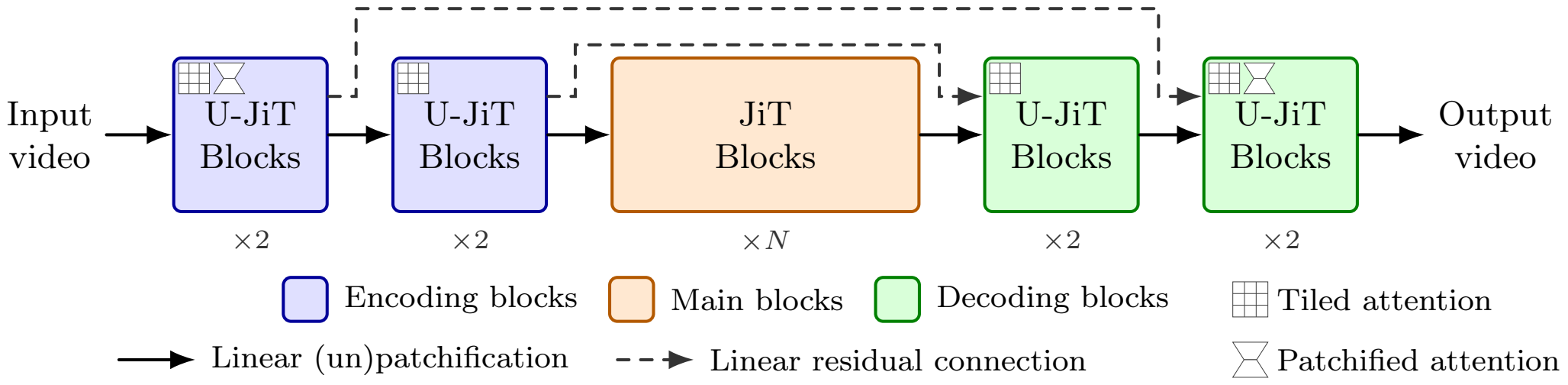
\begin{figure}[t]
    \centering
    \scalebox{0.95}{\begin{tikzpicture}[
    >=Latex,
    font=\small,
    block/.style={
        draw, rounded corners=2pt, thick, align=center,
        minimum width=1.4cm, minimum height=1.4cm, inner sep=4pt,
    },
    enc/.style={block, fill=blue!12, draw=blue!60!black},
    bot/.style={block, fill=orange!18, draw=orange!70!black},
    dec/.style={block, fill=green!15, draw=green!50!black},
    io/.style={align=center, font=\small},
    flow/.style={->, thick, line cap=round},
    skip/.style={->, thick, dashed, gray!40!black, line cap=round},
    rep/.style={font=\scriptsize, gray!40!black},
    iconstroke/.style={very thin, draw=black!75},
]

\newcommand{\tiledicon}[2]{%
    \begin{scope}[shift={#1}]
        \def\s{#2}
        \fill[white] (-0.5*\s,-0.5*\s) rectangle (0.5*\s,0.5*\s);
        \draw[iconstroke] (-0.5*\s,-0.5*\s) rectangle (0.5*\s,0.5*\s);
        \draw[iconstroke] (-\s/6,-0.5*\s) -- (-\s/6,0.5*\s);
        \draw[iconstroke] ( \s/6,-0.5*\s) -- ( \s/6,0.5*\s);
        \draw[iconstroke] (-0.5*\s,-\s/6) -- (0.5*\s,-\s/6);
        \draw[iconstroke] (-0.5*\s, \s/6) -- (0.5*\s, \s/6);
    \end{scope}%
}

\newcommand{\patchicon}[2]{%
    \begin{scope}[shift={#1}]
        \def\s{#2}
        \def\hh{0.5*\s}
        \def\hw{0.5*\s}
        \def\nw{0.18*\s}
        \fill[white] (-\hw,\hh) -- (\hw,\hh) -- (\nw,0) -- (-\nw,0) -- cycle;
        \draw[iconstroke] (-\hw,\hh) -- (\hw,\hh) -- (\nw,0) -- (-\nw,0) -- cycle;
        \fill[white] (-\nw,0) -- (\nw,0) -- (\hw,-\hh) -- (-\hw,-\hh) -- cycle;
        \draw[iconstroke] (-\nw,0) -- (\nw,0) -- (\hw,-\hh) -- (-\hw,-\hh) -- cycle;
    \end{scope}%
}

\node[enc] (e1) at (0,0)   {U-JiT\\ Blocks};
\node[enc] (e2) at (2.0,0) {U-JiT\\ Blocks};
\node[bot, minimum width=2.8cm] (b)  at (4.7,0) {JiT\\ Blocks};
\node[dec] (d2) at (7.4,0) {U-JiT\\ Blocks};
\node[dec] (d1) at (9.4,0) {U-JiT\\ Blocks};

\foreach \n in {e1,e2,d2,d1} {
    \tiledicon{($(\n.north west)+(0.20,-0.20)$)}{0.28}
}
\patchicon{($(e1.north west)+(0.52,-0.20)$)}{0.28}
\patchicon{($(d1.north west)+(0.52,-0.20)$)}{0.28}

\node[rep, below=1pt of e1] {$\times 2$};
\node[rep, below=1pt of e2] {$\times 2$};
\node[rep, below=1pt of b]  {$\times N$};
\node[rep, below=1pt of d2] {$\times 2$};
\node[rep, below=1pt of d1] {$\times 2$};

\node[io, left=0.6cm of e1] (in)  {Input\\ video};
\node[io, right=0.6cm of d1] (out) {Output\\ video};

\draw[flow] (in)  -- (e1);
\draw[flow] (e1)  -- (e2);
\draw[flow] (e2)  -- (b);
\draw[flow] (b)   -- (d2);
\draw[flow] (d2)  -- (d1);
\draw[flow] (d1)  -- (out);

\def\skipstub{0.25}
\def\skipdstub{0.40}
\def\skipyoff{0.35}
\def\skipinner{0.82}
\def\skipouter{1.15}

\draw[skip]
    ($(e2.east)+(0,\skipyoff)$) -- ++(\skipstub,0)
        |- ($(d2.west)+(-\skipdstub,\skipinner)$)
        -- ($(d2.west)+(-\skipdstub,\skipyoff)$)
        -- ($(d2.west)+(0,\skipyoff)$);

\draw[skip]
    ($(e1.east)+(0,\skipyoff)$) -- ++(\skipstub,0)
        |- ($(d1.west)+(-\skipdstub,\skipouter)$)
        -- ($(d1.west)+(-\skipdstub,\skipyoff)$)
        -- ($(d1.west)+(0,\skipyoff)$);

\begin{scope}[shift={(0,-1.5)}]
    \def\lgs{0.4}
    \tikzset{
        legitem/.style={anchor=west, font=\footnotesize, inner sep=1pt},
        legsw/.style={minimum width=\lgs cm, minimum height=\lgs cm, inner sep=0pt},
    }

    \node[enc, legsw] (lenc) at (0.5,0) {};
    \node[legitem] (lenctxt) at ($(lenc.east)+(0.1,0)$) {Encoding blocks};

    \node[bot, legsw] (lbot) at ($(lenctxt.east)+(0.45,0)$) {};
    \node[legitem] (lbottxt) at ($(lbot.east)+(0.1,0)$) {Main blocks};

    \node[dec, legsw] (ldec) at ($(lbottxt.east)+(0.45,0)$) {};
    \node[legitem] (ldectxt) at ($(ldec.east)+(0.1,0)$) {Decoding blocks};

    \begin{scope}[yshift=-0.55cm, xshift=-1.6cm]
        \draw[flow] (0.4,0) -- ++(0.7,0) coordinate (lflowend);
        \node[legitem] (lflowtxt) at ($(lflowend)+(0.1,0)$) {Linear (un)patchification};

        \draw[skip] ($(lflowtxt.east)+(0.4,0)$) -- ++(0.7,0) coordinate (lskipend);
        \node[legitem] at ($(lskipend)+(0.1,0)$) {Linear residual connection};
    \end{scope}

    \coordinate (lattcol) at ($(ldectxt.east)+(0.7,0)$);
    \tiledicon{(lattcol)}{0.32}
    \node[legitem] at ($(lattcol)+(0.20,0)$) {Tiled attention};
    \patchicon{($(lattcol)+(0,-0.55)$)}{0.32}
    \node[legitem] at ($(lattcol)+(0.20,-0.55)$) {Patchified attention};
\end{scope}

\end{tikzpicture}}
    \caption{
        {\bf \ujit{} architecture} overview. Encoding and decoding blocks are parametrized to match standard \jit{} blocks used in the main backbone in computational cost.
    }
    \label{fig:ujit_arch}
    \vspace{-1em}
\end{figure}

\subsection{\ujit{}: varying patch-size transformer}
\label{subsec:method_architecture}

\begin{wrapfigure}{r}{0.4\columnwidth}
    \vspace{-\intextsep}
    \centering
    \scalebox{0.75}{\begin{tikzpicture}[
    >=Latex,
    font=\small,
    block/.style={
        draw, rounded corners=2pt, thick, align=center,
        minimum width=3.0cm, minimum height=0.55cm, inner sep=3pt,
    },
    norm/.style={block, fill=gray!12, draw=gray!60!black},
    op/.style={block, fill=blue!12, draw=blue!60!black},
    mlpop/.style={op, minimum width=1.8cm},
    optop/.style={block, fill=green!15, draw=green!50!black, dashed},
    tileop/.style={block, fill=green!15, draw=green!50!black, dashed, sharp corners},
    mod/.style={block, fill=orange!18, draw=orange!70!black},
    gate/.style={mod},
    sum/.style={draw, circle, thick, inner sep=0pt, minimum size=0.42cm,
                fill=white, font=\small},
    tap/.style={circle, fill=black, inner sep=0pt, minimum size=3pt},
    iconstroke/.style={very thin, draw=black!75},
    flow/.style={thick, line cap=round},
    flowend/.style={->, thick, line cap=round},
    res/.style={->, thick, line cap=round},
    condflow/.style={->, thick, dashed, line cap=round},
    condbus/.style={thick, dashed, line cap=round},
    iolabel/.style={align=center, font=\small},
]

\newcommand{\tiledicon}[2]{%
    \begin{scope}[shift={#1}]
        \def\s{#2}
        \fill[white] (-0.5*\s,-0.5*\s) rectangle (0.5*\s,0.5*\s);
        \draw[iconstroke] (-0.5*\s,-0.5*\s) rectangle (0.5*\s,0.5*\s);
        \draw[iconstroke] (-\s/6,-0.5*\s) -- (-\s/6,0.5*\s);
        \draw[iconstroke] ( \s/6,-0.5*\s) -- ( \s/6,0.5*\s);
        \draw[iconstroke] (-0.5*\s,-\s/6) -- (0.5*\s,-\s/6);
        \draw[iconstroke] (-0.5*\s, \s/6) -- (0.5*\s, \s/6);
    \end{scope}%
}
\newcommand{\patchdownicon}[2]{%
    \begin{scope}[shift={#1}]
        \def\s{#2}
        \fill[white] (-0.5*\s,-0.5*\s) -- (0.5*\s,-0.5*\s) -- (0.18*\s,0.5*\s) -- (-0.18*\s,0.5*\s) -- cycle;
        \draw[iconstroke] (-0.5*\s,-0.5*\s) -- (0.5*\s,-0.5*\s) -- (0.18*\s,0.5*\s) -- (-0.18*\s,0.5*\s) -- cycle;
    \end{scope}%
}
\newcommand{\patchupicon}[2]{%
    \begin{scope}[shift={#1}]
        \def\s{#2}
        \fill[white] (-0.18*\s,-0.5*\s) -- (0.18*\s,-0.5*\s) -- (0.5*\s,0.5*\s) -- (-0.5*\s,0.5*\s) -- cycle;
        \draw[iconstroke] (-0.18*\s,-0.5*\s) -- (0.18*\s,-0.5*\s) -- (0.5*\s,0.5*\s) -- (-0.5*\s,0.5*\s) -- cycle;
    \end{scope}%
}

\newcommand{\trapblock}[4]{%
    \node[minimum width=3.0cm, minimum height=0.55cm, inner sep=0pt] (#1) at (\colx, #2*\dy) {};%
    \ifnum#3=1\relax
        \draw[fill=green!15, draw=green!50!black, dashed, thick, rounded corners=2pt]
            (#1.south west) -- (#1.south east) --
            ([xshift=-0.4cm]#1.north east) -- ([xshift=0.4cm]#1.north west) -- cycle;
    \else
        \draw[fill=green!15, draw=green!50!black, dashed, thick, rounded corners=2pt]
            ([xshift=0.4cm]#1.south west) -- ([xshift=-0.4cm]#1.south east) --
            (#1.north east) -- (#1.north west) -- cycle;
    \fi
    \node at (#1) {#4};%
}

\def\dy{0.78}
\def\colx{0}
\def\condx{2.833}
\def\resoff{1.85}

\node[iolabel] (inx)      at (\colx,  0)      {Input tokens};
\node[norm]    (n1)       at (\colx,  1*\dy) {RMSNorm};
\node[mod]     (m1)       at (\colx,  2*\dy) {Scale, Shift};
\trapblock{apatch}{3}{1}{Attn-patchify}
\node[tileop]  (atile)    at (\colx,  4*\dy) {};
\begin{scope}
    \clip[sharp corners] (atile.south west) rectangle (atile.north east);
    \foreach \i in {-3,-2,-1,0,1,2,3} {
        \draw[very thin, green!50!black] ($(atile.south)+(\i*0.375,0)$) -- ($(atile.north)+(\i*0.375,0)$);
    }
    \draw[very thin, green!50!black] (atile.west) -- (atile.east);
\end{scope}
\node at (atile) {\setlength{\fboxsep}{1pt}\colorbox{green!15}{Attn-tiling}};
\node[op]      (att)      at (\colx,  5*\dy) {Self-Attention};
\trapblock{aunpatch}{6}{-1}{Attn-unpatchify}
\node[gate]    (g1)       at (\colx,  7*\dy) {Scale};
\node[sum]     (p1)       at (\colx,  8*\dy) {$+$};
\node[norm]    (n2)       at (\colx,  9*\dy) {RMSNorm};
\node[mod]     (m2)       at (\colx, 10*\dy) {Scale, Shift};
\node[op]      (ffn)      at (\colx, 11*\dy) {FFN};
\node[gate]    (g2)       at (\colx, 12*\dy) {Scale};
\node[sum]     (p2)       at (\colx, 13*\dy) {$+$};
\node[iolabel] (out)      at (\colx, 14*\dy) {Output tokens};

\draw[flow] (inx)      -- (n1);
\draw[flow] (n1)       -- (m1);
\draw[flow] (m1)       -- (apatch);
\draw[flow] (apatch)   -- (atile);
\draw[flow] (atile)    -- (att);
\draw[flow] (att)      -- (aunpatch);
\draw[flow] (aunpatch) -- (g1);
\draw[flow] (g1)       -- (p1);
\draw[flow] (p1)       -- (n2);
\draw[flow] (n2)       -- (m2);
\draw[flow] (m2)       -- (ffn);
\draw[flow] (ffn)      -- (g2);
\draw[flow] (g2)       -- (p2);
\draw[flowend] (p2)    -- (out);

\coordinate (tap1) at ($(inx.north)!0.55!(n1.south)$);
\coordinate (tap2) at ($(p1.north)!0.55!(n2.south)$);
\node[tap] at (tap1) {};
\node[tap] at (tap2) {};
\draw[res] (tap1) -- ++(-\resoff,0) |- (p1.west);
\draw[res] (tap2) -- ++(-\resoff,0) |- (p2.west);

\node[mlpop]   (mlp)  at (\condx, 1*\dy)   {MLP};
\node[iolabel] (clab) at (\condx, 0) {Class $+$ time};
\draw[flow] (clab) -- (mlp);

\coordinate (busstart) at (mlp.north);
\coordinate (busend)   at (\condx, 12*\dy);
\draw[condbus] (busstart) -- (busend);

\foreach \target in {m1, g1, m2, g2} {
    \draw[condflow] (\target.east -| busstart) -- (\target.east);
}

\end{tikzpicture}}
    \figmidspace
    \caption{
        {\bf \ujit{} block, with attention tiling and patchification blocks.}
    }
    \label{fig:ujit_block}
    \vspace{-1.5em}
\end{wrapfigure}
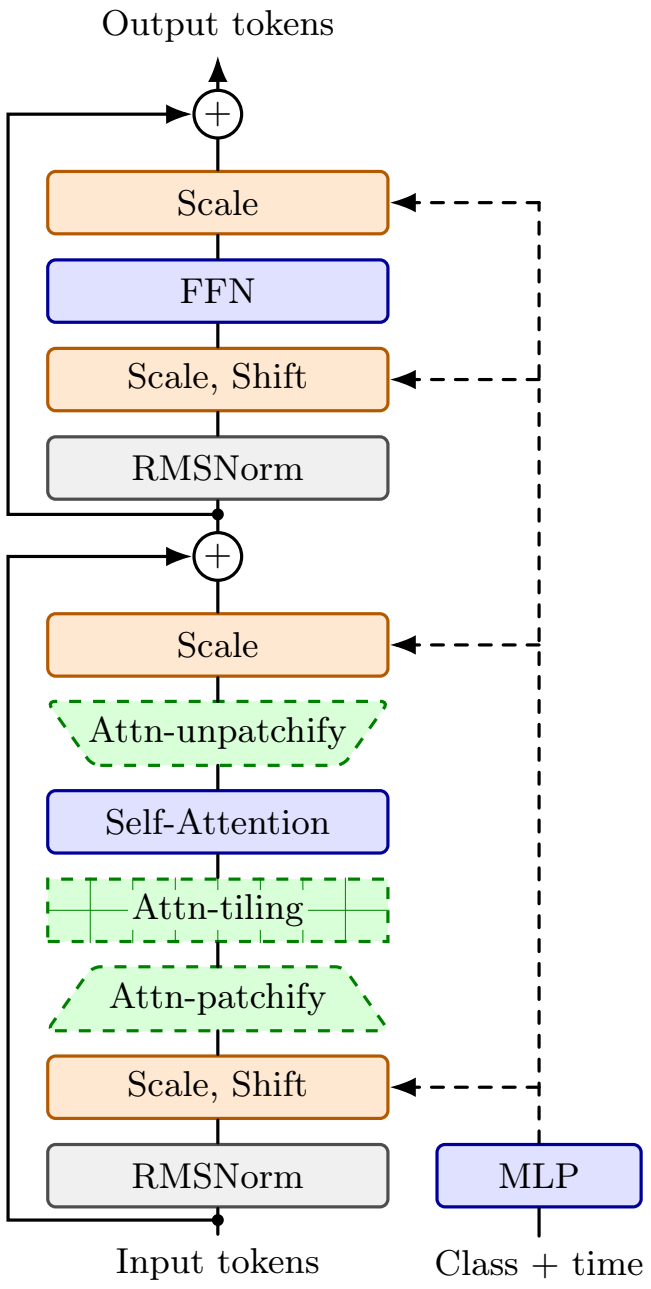

In order to support efficient video generation, a transformer model needs to handle a large spatiotemporal patch size, which reduces the number of tokens.
The analysis in \Cref{subsec:method_bottleneck} motivates  architectural changes to natively handle large patch sizes in pixel-space models.
Existing architectures from the image generation literature~\citep{ddt_Wang_2026_CVPR,pixeldit_Yu_2026_CVPR,Ma_2026_CVPR,Chen_2026_CVPR} address this with a specialized output head, but suffer from two issues:
(1) Scalability: due to the large embedding dimension~\citep{ddt_Wang_2026_CVPR} or large, repeated, conditioning projection~\citep{pixeldit_Yu_2026_CVPR}, their FLOPs and parameter count increase quadratically with the patch size, which makes them impractical for the stronger  compression needed for video generation;
(2) Expressivity: they use a linear projection layer to compress the  input, which is significantly less expressive than the multi-layer encoders employed by latent space methods to extract compacts latents, thereby losing access to useful signal  throughout the model up until the output head.

We propose instead to design our \ujit as a varying-patch-size, U-Net shaped transformer, based on U-DiT~\citep{udit_NEURIPS2024_5d2e24df}, using \jit blocks~\citep{jit_Li_2026_CVPR} and additional adaptations to attention to remain FLOPs-efficient.
This ensures scalability by keeping standard transformer blocks, with only minimal changes in the early (\textit{encoding}) and late (\textit{decoding}) layers.
The overall    architecture is shown in \Cref{fig:ujit_arch}.
We follow the standard DiT~\citep{Peebles_2023_ICCV} model sizes and use 3 levels (2 encoding / decoding levels, and one main level), adjusting the overall number of blocks to keep the same computational cost as \jit.
The central (main) level uses exactly the same blocks as \jit models, with the same patch size and hidden dimension.
Compared to latent space models, our approach incorporates a small encoder and decoder inside the denoising model, and trains in end-to-end manner.

The encoding and decoding blocks, referred to as \ujit{} blocks (see \Cref{fig:ujit_block}), use a smaller patch size, and scale their hidden dimension relative to their patch size to preserve the same number of FLOPs as \jit blocks.
Similarly to U-DiT, we use a single residual connection per level, projected with a zero-initialized linear layer.
Downsampling and upsampling operations are learned as a linear operation combined with spacetime-to-depth / depth-to-spacetime reshaping, similarly to the input / output projections of \jit models.
Furthermore, to allow patchification of individual frames, this linear projection is expressed as two linear layers, one for the spatial dimension and one for the temporal dimension, with the second one skipped for tokens representing individual frames.
We additionally perform the adaptations detailed below to the attention mechanism of the encoding / decoding blocks to remain FLOP-efficient with small patch sizes on videos.
Ablations on the impact of each architectural component of our model are provided in \Cref{tab:ablations}.

\mypar{Attention tiling} Similarly to U-DiT~\citep{udit_NEURIPS2024_5d2e24df}, we use attention tiling to reduce attention complexity inside encoding / decoding layers, by splitting the token grid without changing the attention operation itself. Instead of fixing the number of grid splits, we use an attention tiling window size of $8\!\times\!8\!\times\!4$ for all our models, which scales linearly with video length.

\mypar{Attention patchification} In order to support (near) atomic  spatiotemporal patches of  size $1\!\times\!1\!\times\!1$ or $2\!\times\!2\!\times\!1$, we use a patchified attention, following the linear compressed attention of PixelDiT~\citep{pixeldit_Yu_2026_CVPR}.
We group tokens into spatiotemporal patches that are linearly projected to obtain a smaller token grid, with a higher embedding dimension per token, before applying the attention. A similar linear expansion is applied to the output of the attention, to recover the original token grid.
In practice, we use attention patchification only for the first encoding / last decoding blocks (pixel-level blocks), with the same compression patch size as the second level of the \ujit{}. We scale the attention hidden dimension to preserve FLOPs.
See \Cref{app:implementation} for complete paramtetrization details.

\subsection{Analyzing the two-stage generative process of latent-space models}
\label{subsec:method_latentspace}

\begin{wrapfigure}{r}{0.45\columnwidth}
    \centering
    \includegraphics[width=\linewidth]{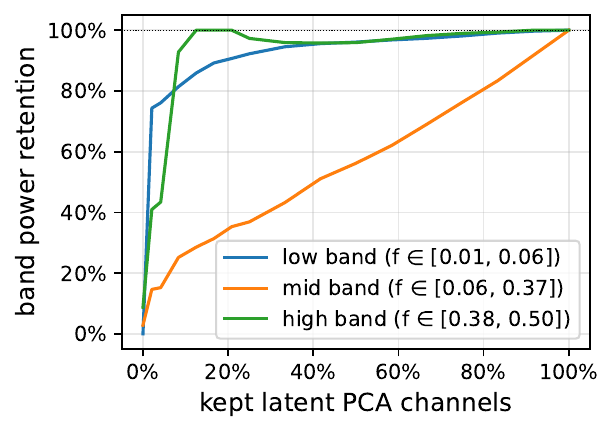}
    \vspace{-1em}
    \caption{
        {\bf Band power retention in decoded videos, depending on latent capacity.}
    }
    \label{fig:band_power}
    \vspace{-1em}
\end{wrapfigure}

Latent space generative models can be seen as a two-stage generative process, where the first stage (latent flow-matching) iteratively generates a low-dimensional latent representation of the video, and the second stage decodes this latent representation into a high-dimensional video as a single decoding step.
We make three key observations that reflect the different behavior of  these two stages.
We support each with analysis on a latent \ljit[B] model trained using the Wan2.2 tokenizer on $256\!\times\!256$ resolution, on 16-frame Kinetics-600 clips.

\begin{wrapfigure}{r}{0.3\columnwidth}
    \vspace{-\intextsep}
    \centering
    \includegraphics[width=\linewidth]{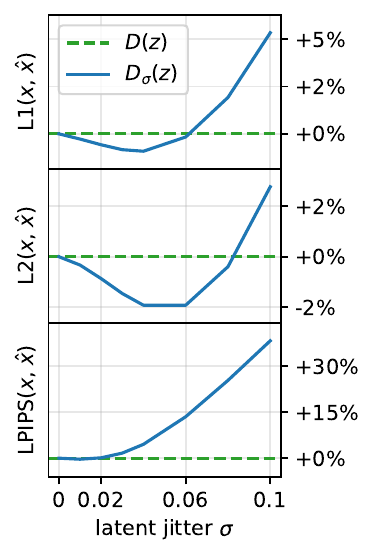}
    \vspace{-1em}
    \caption{
        {\bf Reconstruction quality over latent jittering.}
    }
    \label{fig:averaged_encoder_metrics}
    \vspace{-1.5em}
\end{wrapfigure}

\mypar{1. Low/high frequency split between latent generator and decoder}
Latent-space models are effectively two-stage generative processes.
The latent space is usually significantly lower than that of the video,
which limits the latents to mainly capturing low and mid frequencies, while the decoder is responsible for resolving  high frequency detail.

To identify which frequency bands are captured by the latent and which ones are generated by the decoder,
we reduce the latent capacity by applying a PCA decomposition on it, keeping only the first $k$ PCA components.
We sweep over $k$ and display in \Cref{fig:band_power} the power retention ratio of the low, mid and high frequency bands in the decoded video, compared to the full-latent decoded video.
We observe that low frequencies are mostly captured by the first PCA components (around 90\% power retention with 20\% of the PCA components), while mid frequencies increase almost linearly with the number of PCA components, indicating that successive PCA components capture from low to mid frequencies.
Conversely, high frequencies consistently keep from 90\% to 100\% of their power, even when removing 90\% of the PCA components, at which point the latent encodes almost only low frequencies.
This indicates that the decoder is responsible for generating most of the high frequencies,
while the latent process generates low to mid bands.

\mypar{2. On-manifold high-frequency generation}
Adversarially-trained decoders are optimized to generate on-manifold high-frequency details: from a given latent encoding $z$, the decoder does not generate the most likely video in the ambient space (the closest in MSE on average), but the most likely one on the embedded manifold.
The theoretical insights behind this effect are discussed by \citet{Blau_2018_CVPR}.

To identify if the latent-to-video decoder effectively benefits from this effect, we design a simple variant $D_\sigma$ of the original decoder $D$ to increase ambient space reconstruction fidelity.
Given an input video $x$ and its latent encoding $z$, we reconstruct the video by averaging the output of $D$ over $K=16$ randomly jittered latent encodings, yielding $\hat{x}=D_\sigma(z) = \frac{1}{K} \sum_{k=1}^{K} D(z + \sigma \varepsilon_k)$ where $\varepsilon_k \sim \mathcal{N}(0, I)$.
The reconstruction $\hat{x}$ is a locally smoothed output of the decoder.
In \Cref{fig:averaged_encoder_metrics} we  plot reconstruction metrics, sweeping over the hyperparameter $\sigma$.
For $\sigma\in [0.02, 0.06]$, our jittered decoder pulls the decoded video closer to the initial datapoint, as measured by $L_1$ and $L_2$ distances.
Conversely, the perceptual quality of the video is degraded, as measured by LPIPS.
This confirms that the decoder actively pushes the generated video on-manifold, away from the closest reconstruction in ambient space.

\begin{wrapfigure}{r}{0.3\columnwidth}
    \vspace{-1.5em}
    \centering
    \includegraphics[width=\linewidth]{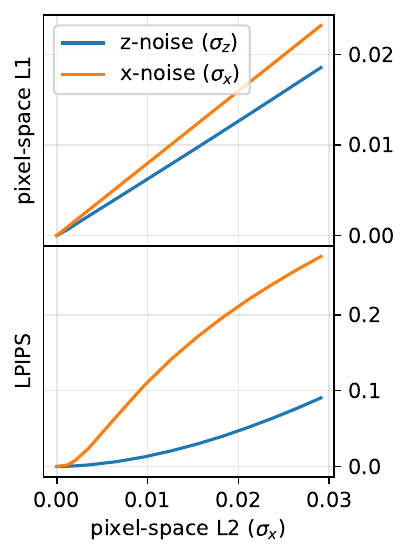}
    \vspace{-1em}
    \caption{
        {\bf Effect of two types of noise on metrics.}
    }
    \label{fig:noise_directionality}
    \vspace{-1.5em}
\end{wrapfigure}

\mypar{3. Perceptually-aligned variance}
The latent Flow-Matching model is trained with an isotropic Gaussian noise schedule and an $L_2$ (MSE) objective, which is agnostic to any perceptual metric and weights every direction in latent space equally.
As a result, its residual error is approximately isotropic in the latent space.

We investigate how the decoder maps isotropic latent variations to anisotropic, perceptually aligned variations in the video pixel space.
To this end we decode a latent $z$ into $\hat{x}$, and a perturbed latent $z + \sigma_z \varepsilon_z$ with isotropic noise into $\hat{x}_{\sigma_z}$.
Averaging across noise and samples,
we measure the effective pixel-space noise scale as the $L_2$ distance $\sigma_x$ between the clean decoded video $\hat{x}$ and the perturbed decoded video $\hat{x}_{\sigma_z}$.
We similarly perturb the reconstructed video $\hat{x}$ directly with isotropic noise in pixel space, as $\hat{x} + \sigma_x \varepsilon_x$.
We compare $L_1$ and LPIPS metrics between both latent ($z$-noise) and pixel-space ($x$-noise) perturbations in \Cref{fig:noise_directionality}.
For a similar, increasing pixel-space noise magnitude, the LPIPS metric deteriorates much faster for the pixel-space perturbation than for the latent perturbation.
This shows that the decoder effectively maps isotropic latent noise to perceptually aligned variations in the video space, which is crucial to generate on-manifold high-frequency details.
Additionally, the $L_1$ distance to the original decoded sample also increases slightly faster, indicating that the latent perturbation is mapped to a sparser variation.

\subsection{Sampling on-manifold in pixel space}
\label{subsec:method_sampling}

\begin{wraptable}{r}{0.4\columnwidth}
    \vspace{-1.5em}
    \centering
    \footnotesize
    \setlength{\tabcolsep}{3pt}
    \caption{
        {\bf \ujit[B] sampling sweep on Kinetics-600, at 200 epochs.}
        A CFG of 1 reverts to non-CFG sampling, and $\sigma_t\!=\!1$ to linear time sampling.
    }
    \label{tab:ut_sampling}
    \resizebox{\linewidth}{!}{%
    \begin{tabular}{c l cccccc}
        \toprule
        & & \multicolumn{6}{c}{CFG} \\
        \cmidrule(lr){3-8}
        & $\sigma_t$ & 1 & 1.5 & 2 & 2.5 & 3 & 3.5 \\
        \midrule
        \multirow{5}{*}{\rotatebox[origin=c]{90}{FVD@5k \odec}}
        & 1  & 3.43 & 3.21 & 3.15 & 3.15 & 3.16 & 3.17 \\
        & 2  & 3.37 & 3.17 & 3.14 & 3.13 & 3.13 & 3.15 \\
        & 4  & 3.33 & 3.14 & 3.12 & 3.11 & 3.12 & 3.15 \\
        & 8  & 3.29 & 3.14 & \textbf{3.10} & 3.14 & 3.20 & 3.29 \\
        & 16 & 3.32 & 3.15 & 3.14 & 3.26 & 3.47 & 3.78 \\
        \midrule
        \multirow{5}{*}{\rotatebox[origin=c]{90}{FID@5k \odec}}
        & 1  & 21.9 & 18.4 & 17.1 & 16.5 & 16.1 & 15.9 \\
        & 2  & 19.0 & 16.3 & 15.4 & 15.0 & 14.9 & 14.9 \\
        & 4  & 17.2 & 15.0 & 14.2 & 13.9 & 14.0 & 14.2 \\
        & 8  & 15.7 & 14.0 & 13.5 & 13.6 & 14.1 & 14.9 \\
        & 16 & 14.9 & \textbf{13.4} & \textbf{13.4} & 14.2 & 15.8 & 18.0 \\
        \bottomrule
    \end{tabular}%
    }
    \vspace{-1.5em}
\end{wraptable}

In \Cref{subsec:method_latentspace} we identified three key mechanisms, specific to latent-space models, that improve the generation of high-quality videos.
Based on this analysis we propose adaptations of  the training and sampling processes of pixel-space models, using known methods from Flow-Matching literature,
to improve pixel-space methods and bridge the gap with  latent-space generation.

\mypar{Time-shifting: focusing  sampling on low-frequencies}
Latent models use most of their compute on the low-frequency structure of the video (latent flow-matching), and a single decoding step to generate high-frequency details.
To mimick this behavior in pixel-space models, we use time-shifting~\citep{pmlr-v202-hoogeboom23a} \textit{at inference only} to shift the sampling process toward low frequencies, keeping only a few steps to generate the high-frequency details.
In practice, given $N$ sampling steps, we use the sampling schedule $t_i=\frac{i}{i + \sigma_t(N - i)}$, reverting to the linear time sampling  schedule for  $\sigma_t=1$.
We validate in a simple setting in \Cref{tab:ut_sampling} that time-shifting can lead to significant improvements in both FVD and FID metrics.

\mypar{Classifier-free guidance: on-manifold high-frequency details}
On the standard task of video completion (such as on the Kinetics dataset), where the strong conditioning is coupled with a GAN-trained decoder, classifier-free guidance is usually deemed unnecessary for latent-space models.
We find in \Cref{tab:ut_sampling} that, for pixel-space models, applying classifier-free guidance~\citep{ho2021classifierfree} with an interval~\citep{NEURIPS2024_dd540e1c} during sampling noticeably improves the quality of generated videos, as it allows the model to generate on-manifold high-frequency details.

\mypar{Perceptual loss: aligning generation variations along perceptually-correct directions}
Flow-Matching models are trained to capture the distribution of real training samples, and to generalize to realistic non-training samples.
To ensure that the variations in high frequencies are aligned with perceptually-correct directions, we apply a perceptual LPIPS loss~\citep{Zhang_2018_CVPR} during training.
Ablations on the impact of this loss  are included in \Cref{tab:ablations}.

We found time-shifting and classifier-free guidance to also improve, to a lesser extent, the performance of latent-space models on all tasks (see \Cref{tab:lat_sampling}).
To ensure a fair comparison, we use them in all our experiments.
We find that a time shift $\sigma_t= 8$ performs best on average across tasks and models for pixel-space models, and $\sigma_t= 4$ for latent-space models.
We select the CFG value for each model on each task with a grid search (step of 0.5), as the optimal CFG value differs substantially across tasks and models.
Following \citep{NEURIPS2024_dd540e1c}, we apply CFG only in the ODE time interval of $[0.1, 1]$ for all models.

\subsection{Model architecture and training details}

We follow \citet{ldit_Yao_2025_CVPR} by using DiT models~\citep{Peebles_2023_ICCV} with the same improvements: SwiGLU activations~\citep{shazeer2020gluvariantsimprovetransformer}, RMSNorm layers~\citep{NEURIPS2019_1e8a1942}, RoPE positional embeddings~\citep{su2024roformer}, and qk-norm~\citep{henry-etal-2020-query}, together with the additional context tokens introduced by \citet{jit_Li_2026_CVPR}.
We refer to these models as \jit{} models in pixel space, \ljit{} models when trained in latent space, and \ujit{} when incorporating our architecture modifications.
We train all models with the same training objective as \jit~\citep{jit_Li_2026_CVPR}, using $x$-prediction and a velocity loss, which we found more effective than $v$-prediction for both pixel-space and latent-space video models.
We use noise scaling to scale across video resolutions and lengths.
We refer to \Cref{app:implementation} for additional implementation details and per-model FLOPs and parametrization.

\section{Experiments}

\subsection{Experimental Setup}
\label{subsec:exp_setup}

We train and evaluate our models on the Kinetics-600 dataset~\citep{carreira2018shortnotekinetics600}, and use the FVD metric~\citep{unterthiner2019accurategenerativemodelsvideo} to evaluate overall video quality.
As FVD assesses distribution match  using spatio-temporal features from an action recognition model, we also evaluate FID~\citep{NIPS2017_8a1d6947} to assess per-frame quality.
We compare our pixel-space models to latent-space models trained using the following state-of-the-art video tokenizers: Wan2.1 and Wan2.2~\citep{wan2025wanopenadvancedlargescale}, LTX~\citep{hacohen2024ltxvideorealtimevideolatent}, and the continuous Cosmos tokenizer~\citep{nvidia2025cosmosworldfoundationmodel}.
We write f$F$t$T$ for a spatiotemporal compression rate of $F$ along the spatial axes and $T$ along the temporal axis.
As the Wan2.2 and LTX video tokenizers have respective compression rates of f16t4 and f32t8, we only use them in strong compression settings.
Additional details about training and evaluation are provided in \Cref{app:implementation}.

\begin{table}[h]
    \centering
    \small
    \setlength{\tabcolsep}{6pt}
    \caption{
        Frames to video: 5 conditioning frames, 12 generated frames, at resolution $128\!\times\!128$ with spatiotemporal compression f8t4.
        B- and XL-sized models.
        FLOPs are computed for a single forward pass of the model, not accounting for the decoder cost of latent models.
        \textit{Pixel} column indicates pixel-space (\checkmark) and latent-space (\(\times\)) generative models.
    }
    \label{tab:5t11_128_XL}
    \begin{tabular}{l c c c | c c}
        \toprule
        Model (f8t4) & Pixel & \#Prms.\ & FLOPs & \odec~FVD  & \odec~FID \\
        \midrule
        Video Diffusion Models~\citep{video_diffusion_ho2022}  & \checkmark & 1.1B & 4.1T & 16.2  & - \\
        RIN~\citep{pmlr-v202-jabri23a}  & \checkmark & 411M & 386G & 10.8  & - \\
        MAGVIT-v2~\citep{yu2024languagemodelbeatsdiffusion}  & $\times$ & 307M & - & 4.3  & - \\
        DeRA~\citep{guo2025deradecoupledrepresentationalignment} & $\times$ & 632M & - & 4.1  & - \\
        W.A.L.T~\citep{walt}  & $\times$ & 313M & - &  3.3  & - \\
        Unified Latents~\citep{heek2026unifiedlatentsultrain}  & $\times$ & - & - & 1.3  & - \\
        WaiT-XL~\citep{pavasovic2026waitsignalsimplefrequencyaware}  & \checkmark & 678M & 1.3T & 0.84 & - \\

        \midrule

        \lcosmos[B]{} \oursampling{} & $\times$ & 124M & 264G & 1.50 & 8.36 \\
        \lwanone[B]{} \oursampling{} & $\times$ & 124M & 264G & 1.09 & 12.13 \\
        \jit[B] \basesamp{}    & \checkmark & 125M & 266G & 1.68 & 45.20 \\
        \jit[B] \oursampling{} & \checkmark & 125M & 266G & 1.22 & 7.28 \\
        \ujit[B]{}~(ours, base sampling)   & \checkmark &  98M & 265G & 1.25 & 7.17 \\
        \ujit[B]{}~(ours)    & \checkmark &  98M & 265G & 1.06 & \textbf{3.08} \\

        \midrule

        \lcosmos[XL]{} \oursampling{} & $\times$ & 643M & 1.3T & 1.00 & 8.78 \\
        \lwanone[XL]{} \oursampling{} & $\times$ & 643M & 1.3T & \underline{0.62} & 11.05 \\

        \jit[XL] \basesamp{}    & \checkmark & 644M & 1.3T & 0.86 & 32.43 \\
        \jit[XL] \oursampling{} & \checkmark & 644M & 1.3T & 0.71 & 5.29 \\

        \ujit[XL]{}~(ours) & \checkmark & 587M & 1.3T & \textbf{0.61} & \underline{3.15} \\
        \bottomrule
    \end{tabular}
\end{table}

\subsection{Frames to video generation}
\label{subsec:frames_to_video}

We evaluate our models on the standard frames-to-video generation task~\citep{video_diffusion_ho2022}, where the model is given 5 frames as input, and is evaluated on generating the next 11 frames (in practice, 12 are generated due to patchification).
We follow recent work~\citep{pavasovic2026waitsignalsimplefrequencyaware,heek2026unifiedlatentsultrain,walt} by training and evaluating at resolution $128\!\times\!128$ with spatiotemporal compression f8t4.
We first train \ljit[B], \jit[B] and \ujit[B] models on this task, and report their results in \Cref{tab:5t11_128_XL}.
Our latent-space model (\ljit[XL] with Wan2.1 tokenizer) outperform existing latent-space and pixel-space video generation methods.
Results show that our \ujit{} is on par with the best latent-space model in terms of FVD, and significantly better in terms of FID.
We also compare \ujit[B] with \jit[B] under both our sampling strategy from \Cref{subsec:method_sampling} and \textit{base sampling} that does not use CFG nor time-shifting.\footnote{For both latent \ljit and pixel-space \jit and \ujit models we otherwise sweep for the best CFG and time shifting, where a values of 1 are equivalent to base sampling. }.
This confirms that both our sampling strategy and architecture each significantly contribute, on their own, to closing the gap between latent-space and pixel-space models, while combining both is necessary to match the FVD surpass the FID of latent-space models.

\mypar{Scaling our models}
To validate the improvements brought by our model at larger scales, we increase the model size to XL on the same task in \Cref{tab:5t11_128_XL}.
We observe clear FVD improvements for all methods, and similar FID for \ujit and \ljit[]. Our model stays on par with the best latent-space model in terms of FVD, with a clear per-frame FID quality improvement, and is noticeably better than \jit[].
We also reach state-of-the-art performance on the Kinetics-600 benchmark, with a $27\%$ improvement from 0.84 to 0.61 in FVD compared to the previous best published results.

\begin{figure}[b]
    \centering
    \includegraphics[width=0.49\linewidth]{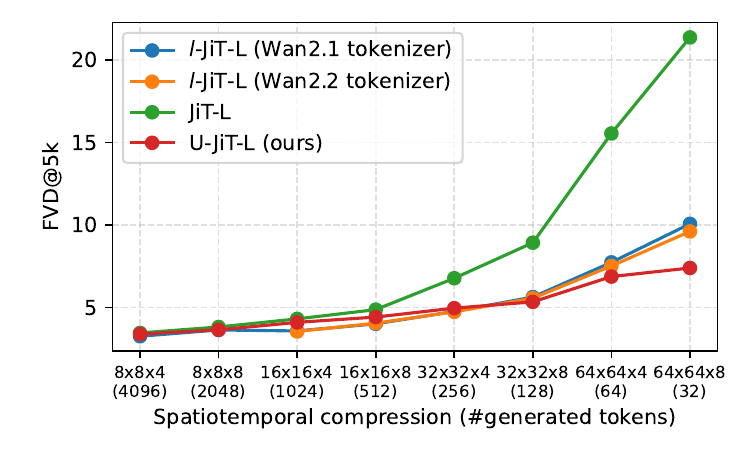}
    \hfill
    \includegraphics[width=0.49\linewidth]{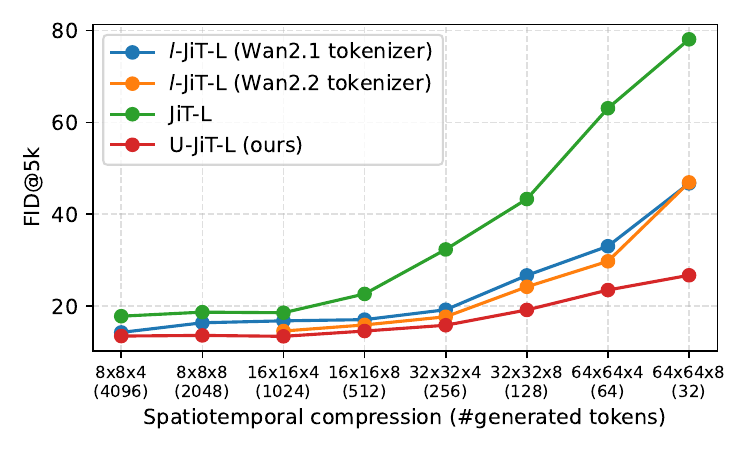}
    \vspace{-0.5em}
    \caption{
        {\bf Scaling the spatiotemporal compression rate.} FVD@5k (left) and FID@5k (right) as functions of the compression. Evaluated at 400 epochs, 1 to 15 frames, resolution $256\!\times\!256$.
    }
    \label{fig:fid_fvd_compression_rate}
\end{figure}

\subsection{Strong video compression}
\label{subsec:scaling_compression_rate}

As established in \Cref{subsec:method_bottleneck}, the main challenge of pixel-space video generation lies in the high dimensionality of the video data, which is exacerbated by the use of strong spatiotemporal compression, necessary for fast, high-resolution generation.
To this end, we study the effect of increased spatiotemporal compression on our models, compared to latent-space and vanilla \jit.

\mypar{Scaling along the compression rate}
We first train \ujit[L], \jit[L] and \ljit[L] (using the Wan2.1 and Wan2.2 tokenizers) models on the single-frame-to-video generation task, at resolution $256\!\times\!256$, where the model is given a single frame as input and is evaluated on the next 15 frames (in practice, 16 are generated and the last one is discarded).
Models are trained for 400 epochs and evaluated on 5k samples.
We report results in \Cref{fig:fid_fvd_compression_rate}.
The \jit{} baseline shows a clear degradation in performance compared to latent-space models, as it struggles with high-dimensional data.
On the other hand, our \ujit{} model is as robust as latent-space models to increased spatiotemporal compression, following closely their FVD, and improving over them for the strongest f64t8 compression setting.
It additionally shows improved per-frame quality (FID) at high compression rates.

\pagebreak
\begin{wraptable}{r}{0.5\columnwidth}
    \centering
    \footnotesize
    \setlength{\tabcolsep}{4pt}
    \caption{
        \textbf{Image to video, strong compression}. 1 to 16 frames, at resolution $256\times256$ with spatiotemporal compression f32t8.
    }
    \label{tab:1t15_256_deep}
    \resizebox{\linewidth}{!}{%
    \begin{tabular}{l c | c c}
        \toprule
        Model (f32t8) & \#Prms.\ & \odec~FVD  & \odec~FID \\
        \midrule
        \lcosmos[XL]{} \oursampling{}        & 644M & 3.94 & 12.41 \\
        \lwanone[XL]{} \oursampling{}        & 644M & 2.73 & 12.86 \\
        \lwantwo[XL]{} \oursampling{}        & 644M & \underline{2.66} & \underline{9.97} \\
        \lltx[XL]{} \oursampling{}           & 643M & 5.09 & 21.06 \\
        \midrule
        \jit[XL]{} \basesamp{}                & 700M & 8.98 & 52.53 \\
        \jit[XL]{} \oursampling{}             & 700M & 5.86 & 30.27 \\
        \ujit[XL]{}~(ours)                    & 478M & \textbf{2.55} & \textbf{7.20} \\
        \bottomrule
    \end{tabular}%
    }
    \vspace{-1em}
\end{wraptable}

\mypar{Strong compression analysis}
We validate our findings by scaling to XL-sized models on the same task at a strong compression rate (f32t8), training until convergence (600 epochs) and evaluating on 50k samples.
Results in \Cref{tab:1t15_256_deep} confirm our patch size scaling analysis, with \ujit[XL] outperforming all latent-space models in terms of both FVD and FID, and \jit[XL] suffering from a clear degradation in performance.

\wrapfill

\subsection{Class to video generation}
\label{subsec:class_to_video}

\begin{wraptable}{r}{0.5\columnwidth}
    \centering
    \footnotesize
    \setlength{\tabcolsep}{4pt}
    \caption{
        \textbf{Class to video}. 65 frames generated from a class condition, at resolution $256\times256$ with strong spatiotemporal compression f32t8.
}
    \label{tab:c2v}
    \resizebox{\linewidth}{!}{%
    \begin{tabular}{l c | c c}
        \toprule
        Model (f32t8) & \#Prms. & \odec~FVD  & \odec~FID \\
        \midrule
        \lcosmos[XL]{} \oursampling{}        & 647M & 4.78 & \underline{17.44} \\
        \lwanone[XL]{} \oursampling{}        & 647M & \underline{3.31} & 21.52 \\
        \lwantwo[XL]{} \oursampling{}        & 647M & 3.55 & 18.08 \\
        \lltx[XL]{} \oursampling{}           & 646M & 3.96 & 22.24 \\
        \midrule
        \jit[XL]{} \basesamp{}                & 707M & 9.45 & 65.49 \\
        \jit[XL]{} \oursampling{}             & 707M & 6.88 & 32.60 \\
        \ujit[XL]{}~(ours)                    & 478M & \textbf{3.00} & \textbf{12.65} \\
        \bottomrule
    \end{tabular}%
    }
\end{wraptable}

Video completion evaluates the ability to continue a sequence of frames, with a strong conditioning signal that specifies most of the structure and background of the video.
This mainly focuses on the capacity of the model to output coherent motion, and high-quality frames similar to the input, which is close to an auto-encoding task.
We also evaluate our method on the harder class-to-video generation task, where the full distribution of the content is learned by the model.
We use the same strong compression setting of f32t8 as in \Cref{subsec:scaling_compression_rate}, training \ljit[XL], \jit[XL] and \ujit[XL] models to generate 65 frames from a class condition, corresponding to 8 seconds of video, at resolution $256\!\times\!256$.
To compute FVD, we split each of the 50k generated videos into 4 clips of 16 frames, yielding a total of 200k clips for evaluation.
Results reported in \Cref{tab:c2v} show that our \ujit outperforms all latent-space models, increasing the gap in FVD improvement from 4\% to 9\% compared to the image-to-video task (\Cref{tab:1t15_256_deep}).
This confirms that \ujit generalizes to general video generation tasks, learning the full distribution of low- and high-level features.

\section{Conclusion}

Two-stage latent-space generation models are inherently limited by the quality of the tokenizer, whose objectives are not easily aligned with the generation task: reconstruction fidelity, compression ratio and diffusability are all competing objectives that are hard to optimize simultaneously.
End-to-end pixel-space generation offers a paradigm that bypasses the tokenizer bottleneck, but is limited by the high dimensionality of the data space.
Previous work has shown how to separate high-dimensional noise from the compressible signal to provide accurate velocity prediction in ambient space.
In this work, we show that this is not sufficient to capture the distribution of high-dimensional data such as videos, and propose a simple adaptation of the transformer architecture to address this issue.
Our findings show that \ujit,  a multi-resolution pixel-space model, can be trained end-to-end to achieve  results competitive with latent-space models, while providing better scalability to strong compression settings.
We hope that this work will encourage further research on end-to-end video generation.

\clearpage
\newpage
\bibliographystyle{assets/plainnat}
\bibliography{bibliography}
\clearpage
\newpage

\appendix
\onecolumn
\clearpage

\setcounter{figure}{0}
\renewcommand\thefigure{S\arabic{figure}}
\setcounter{table}{0}
\renewcommand{\thetable}{S\arabic{table}}

\theoremstyle{plain}
\newtheorem{theorem}{Theorem}[section]
\newtheorem{lemma}[theorem]{Lemma}
\newtheorem{proposition}[theorem]{Proposition}
\newtheorem{corollary}[theorem]{Corollary}
\theoremstyle{definition}
\newtheorem{definition}[theorem]{Definition}
\newtheorem{example}[theorem]{Example}
\theoremstyle{remark}
\newtheorem{remark}[theorem]{Remark}
\newtheorem{note}[theorem]{Note}
\newtheorem{statement}{Statement}[section]

\section{Implementation details}
\label{app:implementation}

We base our implementation on a reproduction of the publicly available JiT codebase\footnote{\url{https://github.com/LTH14/JiT}}~\citep{jit_Li_2026_CVPR}.
All our configurations are available in \Cref{tab:config}.
We apply the following modifications on top of JiT:

\begin{itemize}
    \item \textbf{Register tokens}: similarly to how JiT uses in-context class tokens, we use register tokens in the model.
    On frames-to-video completion, they are a set of learnable embeddings that are concatenated to the input of the transformer blocks.
    On class-to-video generation, they are strictly equivalent to JiT in-context class tokens, as we additionally embed the class label into the register tokens.
    \item \textbf{Bottleneck}: we do not use the bottleneck in the input projection of \jit{} models, unless explicitly specified (preliminary and ablation experiments).
    \item \textbf{Training time sampler}: we use a heavy-tailed training time sampler for pixel-space diffusion, as a mixture of uniform and logit-normal distributions.
    \item \textbf{Noise scaling}: we sweep over noise scaling factors of the form $2^{k}$ for each video resolution and length, and for each individual tokenizer, as we find that the optimal scaling factor varies greatly across both tokenizers, and pixel-space resolutions.
    \item \textbf{Attention implementation}: we use FlashAttention~\citep{NEURIPS2022_67d57c32} for all attention layers to support the large token count of video models.
\end{itemize}

\begin{table}[h]
    \centering
    \small
    \setlength{\tabcolsep}{6pt}
    \caption{
        {\bf Configurations of experiments.}
    }
    \label{tab:config}
    \begin{tabular}{l *{3}{>{\centering\arraybackslash}p{1.8cm}}}
        \toprule
         & \jit[B] & \jit[L] & \jit[XL] \\
        \midrule
        \rowcolor{black!10}\multicolumn{4}{l}{\textbf{architecture}} \\
        depth        & 12  & 24   & 28 \\
        hidden dim   & 768 & 1024 & 1152 \\
        heads        & 12  & 16   & 18 \\
        dropout             & \multicolumn{3}{c}{0} \\
        register tokens & \multicolumn{3}{c}{32} \\
        registers start block & 4 & 8 & 8 \\
        \midrule
        \rowcolor{black!10}\multicolumn{4}{l}{\textbf{training}} \\
        epochs                       & \multicolumn{3}{c}{600 (unless specified otherwise)} \\
        warmup epochs                & \multicolumn{3}{c}{5} \\
        optimizer                    & \multicolumn{3}{c}{AdamW, $\beta_1, \beta_2 = 0.9, 0.95$} \\
        batch size                   & \multicolumn{3}{c}{128} \\
        learning rate                & \multicolumn{3}{c}{2e-4 (\jit{}, \ujit{}), 4e-4 (\ljit{})} \\
        learning rate schedule       & \multicolumn{3}{c}{constant} \\
        weight decay                 & \multicolumn{3}{c}{0.01} \\
        EMA decay                    & \multicolumn{3}{c}{$0.9999$} \\
        time sampler        & \multicolumn{3}{c}{$\begin{cases} t \sim \mathcal{U}(0,1) & \text{w.p.\ } p_l \\ \mathrm{logit}(t) \sim \mathcal{N}(\mu_l, \sigma_l^2) & \text{w.p.\ } 1-p_l \end{cases}$} \\
        \textit{time mean}           & \multicolumn{3}{c}{$\mu_l = -0.8$ (\jit{}, \ujit{}), $\mu_l = 0$ (\ljit{})} \\
        \textit{time variance}       & \multicolumn{3}{c}{$\sigma_l = 0.8$ (\jit{}, \ujit{}), $\sigma_l = 1$ (\ljit{})} \\
        \textit{time heavy tail}           & \multicolumn{3}{c}{$p_l = 0.1$ (\jit{}, \ujit{}), $p_l = 0$ (\ljit{})} \\
        clip of $(1-t)$ in division  & \multicolumn{3}{c}{0.05} \\
        class token drop (for CFG)   & \multicolumn{3}{c}{0.1} \\
        LPIPS loss weight             & \multicolumn{3}{c}{0.25 (\ujit{}), 0 (\jit{}, \ljit{})} \\
        \midrule
        \rowcolor{black!10}\multicolumn{4}{l}{\textbf{sampling}} \\
        ODE solver                   & \multicolumn{3}{c}{Heun~\citep{heun1900neue}} \\
        ODE steps                    & \multicolumn{3}{c}{$N=50$} \\
        time steps                   & \multicolumn{3}{c}{$t_i=\frac{i}{i + \sigma_t(N - i)}$} \\
        CFG scale sweep step        & \multicolumn{3}{c}{$0.5$} \\
        CFG interval                 & \multicolumn{3}{c}{$[0.1, 1]$} \\
        \bottomrule
    \end{tabular}
    \tabfigendspace
\end{table}

\mypar{\ujit[] blocks parametrization}
The \ujit{} encoding and decoding blocks are parametrized by their patch size and hidden dimension, with additional attention patchification (patch and dimension) and tiling parameters.
The dimensions and number of those blocks are adjusted per setting (video resolution, length, main level patch size, and model size) to keep a similar or lower computational cost as matching \jit models.
We provide in \Cref{tab:ujit_blocks} the parametrization of each level of the \ujit{} models for each setting used in our experiments, together with the number of parameters and FLOPs compared to JiT.
Main blocks follow the same parametrization as \jit (\Cref{tab:config}).

\begin{table}[h]
    \centering
    \small
    \setlength{\tabcolsep}{4pt}
    \caption{
        {\bf \ujit{} blocks parametrization.}
        Per-setting configuration of the \ujit{} multi-level blocks across video resolutions, lengths, tokenizers, and model sizes, together with parameter counts and per forward-pass FLOPs compared to \jit{}.
    }
    \label{tab:ujit_blocks}
    \begin{tabular}{l *{4}{>{\centering\arraybackslash}p{1.7cm}}}
        \toprule
        \rowcolor{black!10}\multicolumn{5}{l}{\textbf{Setting}} \\
        Video resolution      & \multicolumn{2}{c}{$128 \times 128$} & \multicolumn{2}{c}{$256 \times 256$} \\
        Video length          & \multicolumn{2}{c}{17 frames} & 17 frames & 65 frames \\
        Tokenizer patch size  & \multicolumn{2}{c}{f8t4 ($8\times 8\times 4$)} & \multicolumn{2}{c}{f32t8 ($32\times 32\times 8$)} \\
        Model size            & B & XL & XL & XL \\
        Pixel-space noise scaling  & \multicolumn{2}{c}{2} & 2 & 4 \\

        \midrule
        \rowcolor{black!10}\multicolumn{5}{l}{\textbf{First level}} \\
        Patch size                & \multicolumn{2}{c}{$1\times 1\times 1$} & \multicolumn{2}{c}{$2\times 2\times 1$} \\
        Depth                     & \multicolumn{4}{c}{2 (encoding) + 2 (decoding)} \\
        Hidden dim                & \multicolumn{2}{c}{32} & \multicolumn{2}{c}{24} \\
        Attention patchification  & \multicolumn{2}{c}{$4\times 4\times 1$} & \multicolumn{2}{c}{$8\times 8\times 2$} \\
        Attention dim             & 128 & 384 & 128 & 64 \\
        Attention tiling          & \multicolumn{4}{c}{$8\times 8\times 4$} \\
        \midrule
        \rowcolor{black!10}\multicolumn{5}{l}{\textbf{Second level}} \\
        Patch size                & \multicolumn{2}{c}{$8\times 8\times 1$} & \multicolumn{2}{c}{$8\times 8\times 2$} \\
        Depth                     & \multicolumn{4}{c}{2 (encoding) + 2 (decoding)} \\
        Hidden dim                & 256 & 384 & \multicolumn{2}{c}{128} \\
        Attention tiling          & \multicolumn{4}{c}{$8\times 8\times 4$} \\
        \midrule
        \rowcolor{black!10}\multicolumn{5}{l}{\textbf{Main level}} \\
        Depth                     & 8 & 20 & \multicolumn{2}{c}{20} \\
        \midrule
        \rowcolor{black!10}\multicolumn{5}{l}{\textbf{Efficiency}} \\
        \jit{} parameters   & 124.8M & 644.5M & 700.0M & 706.5M \\
        \ujit{} parameters  & 98.2M  & 587.5M & 478.1M & 478.2M \\
        \jit{} FLOPs        & 266.3G & 1.3T   & 199.1G & 596.1G \\
        \ujit{} FLOPs       & 264.8G & 1.3T   & 195.2G & 575.2G \\
        \bottomrule
    \end{tabular}
    \tabfigendspace
\end{table}

\mypar{First frame embedding}
A standard practice in state-of-the-art video tokenizers~\citep{nvidia2025cosmosworldfoundationmodel,hacohen2024ltxvideorealtimevideolatent,wan2025wanopenadvancedlargescale} is to embed the first frame of the video independently, and use temporal patchification on the remaining frames.
Through the video tokenizers, our latent models implicitly use this single-frame embedding.
We follow this practice for our pixel-space models: for \ujit, each spatiotemporal linear downsampling and upsampling projection (as input, output and between each level) is decomposed as a spatial projection followed by a temporal projection.
On the first frame, the temporal projections are skipped.
For \jit, we use a separate linear projection for the first frame, and a spatiotemporal projection for the remaining frames.

\mypar{Dataset}
We train on the Kinetics-600 dataset~\citep{carreira2018shortnotekinetics600}, subsampling each video to 8 FPS, resizing the shorter side to the target resolution, and using a center crop.
During training, we apply data augmentation by randomly flipping the video horizontally with probability 0.5.
We filter out videos shorter than the target length, depending on the task.

\mypar{Evaluation}
Following common practices, we evaluate FVD~\citep{unterthiner2019accurategenerativemodelsvideo} for video quality and FID~\citep{NIPS2017_8a1d6947} for per-frame quality.
For the FVD evaluation, we use the pytorch adaptation\footnote{\url{https://github.com/piergiaj/pytorch-i3d}} of I3D models trained on the Kinetics dataset~\citep{Carreira_2017_CVPR}.
FID uses the commonly-used torch-fidelity implementation\footnote{\url{https://github.com/toshas/torch-fidelity}}.
Evaluations are conducted on the eval set for frames-to-video completion, and on the training set for class-to-video generation.
We use the first 16 frames of the output (conditioning + generated frames) for FVD evaluation, and the last generated frame for FID.
Both metrics are evaluated on 50k samples.
FVD@5k and FID@5k indicate evaluation on 5k samples instead.

\section{Additional results and analysis}
\label{app:additional_results}

\mypar{Sampling latent-space models}
In \Cref{subsec:method_sampling} we consider sampling methods to fruther bridge the gap between latent-space and pixel-space flow-matching models.
While we observe in \Cref{subsec:method_latentspace} that latent models focus the generative process on lower frequencies and push the generated high frequencies on-manifold through adversarial and perceptual losses, it is not clear whether their action is sufficient for faithful video generation.
We therefore apply both sampling methods (classifier-free guidance and time-shifting) to latent-space models.
In \Cref{tab:lat_sampling}, we study their effect on \lwantwo[B], and observe an improvement on both the FID and FVD metrics when combining the two methods.
The improvement is smaller than the one observed on pixel-space models: FVD and FID improve respectively by 5\% and 28\% when applied to latents, compared to 10\% and 39\% for pixel space.
In our main experiments we therefore report results using the best time-shifting and classifier-free guidance settings for latent-space \ljit models.

\begin{table}[h]
    \centering
    \small
    \setlength{\tabcolsep}{4pt}
    \caption{
        {\bf \lwantwo[B] latent sampling sweep on Kinetics-600 (epoch 200), on 5-to-11 frames video generation.}
        FVD@5k and FID@5k over $\sigma_t$ (time shift) and CFG.
        Lower is better; best per metric in \textbf{bold}.
    }
    \label{tab:lat_sampling}
    \begin{tabular}{c l cccccc}
        \toprule
        & & \multicolumn{6}{c}{CFG} \\
        \cmidrule(lr){3-8}
        & $\sigma_t$ & 1 & 1.5 & 2 & 2.5 & 3 & 3.5 \\
        \midrule
        \multirow{5}{*}{\rotatebox[origin=c]{90}{FVD@5k \odec}}
        & 1  & 3.17 & 3.02 & 3.11 & 3.20 & 3.25 & 3.27 \\
        & 2  & 3.13 & 3.02 & 3.12 & 3.21 & 3.26 & 3.28 \\
        & 4  & 3.08 & \textbf{3.01} & 3.13 & 3.22 & 3.28 & 3.29 \\
        & 8  & 3.05 & 3.01 & 3.15 & 3.22 & 3.29 & 3.34 \\
        & 16 & 3.13 & 3.10 & 3.25 & 3.39 & 3.58 & 3.82 \\
        \midrule
        \multirow{5}{*}{\rotatebox[origin=c]{90}{FID@5k \odec}}
        & 1  & 19.1 & 15.5 & 14.7 & 14.4 & 14.4 & 14.4 \\
        & 2  & 18.5 & 15.1 & 14.3 & 14.1 & 14.0 & 14.1 \\
        & 4  & 17.9 & 14.8 & 14.0 & \textbf{13.8} & \textbf{13.8} & 13.9 \\
        & 8  & 18.3 & 15.2 & 14.5 & 14.4 & 14.7 & 15.3 \\
        & 16 & 21.6 & 17.2 & 16.9 & 18.3 & 20.8 & 24.6 \\
        \bottomrule
    \end{tabular}
    \tabfigendspace
\end{table}

\mypar{Architectural and sampling ablations}
To study the impact of each component of our method, we perform a series of architectural and sampling-time ablations on our model.
We train variants of the \ujit[L] model on the 1-to-15 frame prediction task, with a spatiotemporal compression rate of f32t8, as in \Cref{tab:1t15_256_deep}, and report results in \Cref{tab:ablations}. Line (1) is our default configuration used in our main experiments.

\newcommand{\dsame}{\multicolumn{1}{@{\hspace{4pt}}c|}{\vsame{--}}}

\newcommand{\subrule}{\arrayrulecolor{black!30}\cmidrule{2-8}\arrayrulecolor{black}}

\begin{table}[h]
    \centering
    \small
    \setlength{\tabcolsep}{6pt}
    \caption{
        {\bf Architectural and sampling ablations.}
        Variants of \ujit[L]{} trained for 400 epochs on 1 conditioning frame to 15 evaluated frames, at resolution $256\!\times\!256$ with spatiotemporal compression f32t8.
        The first block ablates the loss, the second the architectural choices (attention adaptations, patchification levels and network structure), and the third the sampling-time methods.
        FLOPs are computed for a single forward pass of the model.
    }
    \label{tab:ablations}
    \begin{tabular}{@{}r@{~}l r@{\hspace{4pt}}l | r@{\hspace{4pt}}l r@{\hspace{4pt}}l}
        \toprule
        & Model (f32t8) & \multicolumn{2}{c|}{FLOPs} & \multicolumn{2}{c}{FVD@5k} & \multicolumn{2}{c}{FID@5k} \\
        \midrule
        \multicolumn{8} {c}{\emph{Default configuration}}\\
        (1)  & \ujit[L]{}~(ours)                  & 137G &                   & 5.35 &                  & 19.20 &                 \\
        \midrule
        \multicolumn{8} {c}{\emph{Loss ablations}}\\
        (2)  & ~~w/o LPIPS                      & 137G & \dsame             & 6.33 & \vworse{$+$0.98}  & 35.47 & \vworse{$+$16.27} \\
        \midrule
        \multicolumn{8} {c}{\emph{Architecture  ablations}}\\
        (3)  & ~~w/o attention tiling           & 358G & \vworse{$+$221G}   & 5.32 & \vbetterl{$-$0.03} & 19.13 & \vbetterl{$-$0.07} \\
        (4)  & ~~w/o attention patchification   & 149G & \vworse{$+$12G}    & 5.63 & \vworsel{$+$0.28}  & 18.38 & \vbetterl{$-$0.82} \\
        \subrule
        (5)  & ~~w/ 2 \ujit{} levels only         & 130G & \vbetterl{$-$7G}    & 5.44 & \vworsel{$+$0.09}  & 19.65 & \vworsel{$+$0.45}  \\
        (6)  & ~~w/ main level only (\jit[L]{})   & 137G & \dsame             & 8.98 & \vworse{$+$3.63}  & 43.33 & \vworse{$+$24.13} \\
        (7)  & ~~w/ main level only and bottleneck & 132G & \vbetterl{$-$5G}    & 10.08 & \vworse{$+$4.73} & 42.37 & \vworse{$+$23.17} \\
        \subrule
        (8)  & ~~w/o residual connections       & 137G & \dsame             & 6.09 & \vworse{$+$0.74}  & 20.01 & \vworsel{$+$0.81}  \\
        (9)  & ~~w/o \ujit{} blocks             & 131G & \vbetterl{$-$6G}    & 7.47 & \vworse{$+$2.12}  & 23.33 & \vworse{$+$4.13}  \\
        \midrule
        \multicolumn{8} {c}{\emph{Sampling ablations}}\\
        (10) & ~~w/o time-shift                 & 137G & \dsame             & 5.49 & \vworsel{$+$0.14}  & 27.59 & \vworse{$+$8.39}  \\
        (11) & ~~w/o CFG                        & 137G & \dsame             & 5.86 & \vworse{$+$0.51}  & 21.09 & \vworse{$+$1.89}  \\
        (12) & ~~w/o time-shift and w/o CFG       & 137G & \dsame             & 6.43 & \vworse{$+$1.08}  & 34.74 & \vworse{$+$15.54} \\
        \bottomrule
    \end{tabular}
    \tabfigendspace
\end{table}

First, when we compare our main model to line 2 we observe that LPIPS loss contributes \textit{significantly} to per-frame quality (FID), while also improving the overall video quality (FVD).

Secondly, we analyze the impact of several of our architectural design choices.
Attention tiling in the encoding and decoding blocks (see line 3) is \textit{necessary} to keep the model efficient and to prevent the attention cost from exploding and more than doubling the total number of FLOPs.
Enabling it only slightly degrades both metrics.
Attention patchification (line 4) provides a 5\% improvement in FVD while degrading FID by 5\%, and slightly reduces the computational cost: while it slightly improves our model overall, this component is \textit{dispensable}.
Next, we remove either one (line 5) or both (line 6) of the encoding-decoding levels of our \ujit architecture, keeping the overall number of layers the same by increasing the main level depth.
Removing the pixel-level (first level) layers (line 5) only causes a small degradation of the metrics.
On the other hand, keeping only the main level (line 6), which results in a \jit[L] model, deteriorates generation quality significantly.
We additionally validate our choice of removing the input bottleneck (line 7) from \jit, as it further degrades both metrics on video generation.
We conclude that the progressive compression from pixel space to the internal representation is \textit{central} to our method, while the exact number of levels and patch sizes are hyperparameters with lower impact.
These ablations  validate that our U-shaped design is key to our method, while attention adaptations mostly focus on enabling efficient implementation of this architecture.

In lines  8 and 9 study the impact of the two components of the U-shaped architecture: residual connections, and progressive patchification.
In line 8 we keep the varying patch size across blocks without the residuals, while in line 9 we use a plain single-resolution transformer (as in \jit), but adding the residual connections.
We observe that while both components contribute to improving FVD and FID metrics, the progressive patchification and unpatchification of U-JiT blocks is the most significant one.

Finally, we study  sampling variants in this setting, complementing the results from \Cref{subsec:method_sampling}.
While both classifier-free guidance and time-shifting impact FVD and FID, removing the time shift mostly has most impact on the  per-frame quality (FID), while removing CFG has most impact on the overall video quality (FVD). Removing both leads to the largest degradations in both metrics, suggesting that time shifting and CFG  act as complementary mechanisms.

\end{document}